\documentclass[conference]{IEEEtran}
\IEEEoverridecommandlockouts
\usepackage{cite}
\usepackage{amsmath,amssymb,amsfonts}
\usepackage{algorithmic}
\usepackage{graphicx}
\usepackage{textcomp}
\usepackage{xcolor}
\usepackage{bbding}
\usepackage{multirow}
\usepackage{algorithm}
\usepackage{algorithmic}
\usepackage{booktabs}
\def\BibTeX{{\rm B\kern-.05em{\sc i\kern-.025em b}\kern-.08em
    T\kern-.1667em\lower.7ex\hbox{E}\kern-.125emX}}

\usepackage[left=0.68in,right=0.68in,top=0.75in,bottom=1.1in]{geometry}

\begin{document}

\title{An Effective Approach for Multi-label Classification with Missing Labels\\
%\thanks{The authors gratefully acknowledge the partial financial support of the
%National Science Foundation (1830512 and 2018966) .}
}

\author{\IEEEauthorblockN{1\textsuperscript{st} Xin Zhang}
\IEEEauthorblockA{%\textit{Electrical Engineering} \\
\textit{University of South Carolina}\\
Columbia, United States \\
xz8@email.sc.edu}
\and
\IEEEauthorblockN{2\textsuperscript{nd} Rabab Abdelfattah}
\IEEEauthorblockA{%\textit{Electrical Engineering} \\
\textit{University of South Carolina}\\
Columbia, United States \\
rabab@email.sc.edu}
\and
\IEEEauthorblockN{3\textsuperscript{rd} Yuqi Song}
\IEEEauthorblockA{%\textit{Computer Science Engineering} \\
\textit{University of South Carolina}\\
Columbia, United States \\
yuqis@email.sc.edu}
\and
\IEEEauthorblockN{4\textsuperscript{th} Xiaofeng Wang}
\IEEEauthorblockA{%\textit{Electrical Engineering} \\
\textit{University of South Carolina}\\
Columbia, United States \\
wangxi@cec.sc.edu}
}
\maketitle

\begin{abstract}
%Classification is the most common problem in deep learning. 
Compared with multi-class classification, multi-label classification that contains more than one class is more suitable in real life scenarios. Obtaining fully labeled high-quality datasets for multi-label classification problems, however, is extremely expensive, and sometimes even infeasible, with respect to annotation efforts, especially when the label spaces are too large.  This motivates the research on partial-label classification, where only a limited number of labels are annotated and the others are missing.  
%
%Otherwise, noisy labels may lead to a significant performance degradation.
%Furthermore, training with fully-labeled datasets containing noisy labels will lead to severely degradation of performance. 
%Furthermore, the imbalance between positives and negatives hinder the performance of classifiers.  
To address this problem, we first propose a pseudo-label based approach to reduce the cost of annotation without bringing additional complexity to the existing classification networks.
Then we quantitatively study the impact of missing labels on the performance of classifier.  Furthermore, by designing a novel loss function, we are able to relax the requirement that each instance must contain at least one positive label, which is commonly used in most existing approaches.
Through comprehensive experiments on three large-scale multi-label image datasets, i.e. MS-COCO, NUS-WIDE, and Pascal VOC12,
we show that our method can handle the imbalance between positive labels and negative labels, while still outperforming existing missing-label learning approaches in most cases, and in some cases even approaches with fully labeled datasets.

%To address this problem, we first propose a pseudo-label based approach to train multi-label classifier with missing-label datasets by designing new loss functions and training schemes, which can effectively reduce the cost of annotation. 
%Then, we systematically study the the performance differences of almost all possible missing-label settings, including partial observed labels (POL), partial positive/ negative labels (PPL/ PNL) and two extreme cases: single positive/ negative label (SPL/ SNL). 
%By comprehensive experiments on three large-scale multi-label image datasets, i.e. MS-COCO, NUS-WIDE and Pascal VOC12,
%we further demonstrate that our method can not only be effectively applied to all the above settings, 
%but also is comparable to existing missing-label learning approaches, and even approaches to the accuracy trained with full labels in some settings.
%Finally, for alleviating the imbalance between positive labels and negative labels, we quantify the impact of negative labels on performance and propose a new scheme by exploiting the label proportion.
%提出一种方法，能够有效解决missing label问题
%量化研究数据缺失程度对performance的影响。特别是对仅包含negative标签的POL setting进行了讨论
%大量实验

%SOTA
\end{abstract}


%不改变the design of network structures, only focus on training schemes and the potential of loss function. which will not increase the complexity of implementation and time consumed by training process
\section{Introduction}
%multi-label问题重要性
In deep learning, multi-class classification is a common problem where the goal is to
classify a set of instances, each associated with a unique class label from a set of disjoint class labels. 
%Deep Convolutional Neural Networks (CNNs) have provided great performance for solving this problem. 
A generalized version of multi-class problem is multi-label classification~\cite{tsoumakas2007multi},
which allows the instances to be associated with more than one class.  It is more a practical problem in real life because of the intrinsic multi-label property of the physical world~\cite{durand2019learning}: 
automatic driving always needs to identify which objects are contained in the current scene, such as cars, traffic lights and pedestrians;
%the basis of vision-based autonomous driving is to examine which objects are contained in the current scene; 
CT scan can detect a variety of possible lesions; a movie can simultaneously belong to different categories, for instance.
%multi label难点在于数据集标注
Ideally, multi-label classification is a form of supervised learning~\cite{hastie2009overview}, which requires lots of accurate labels.  In practice, however, annotating all labels for each training instance raises a great challenge in multi-label classification, which is time-consuming and even impractical especially in the presence of a large number of categories~\cite{cole2021multi,zhang2021simple}. 
Therefore, how to leverage the performance of multi-label classifier and the cost of collecting labels receives significant interests in recent years.  

The main strategies can be roughly divided into two categories: (1) generating annotations automatically and (2) training with missing labels.  The former uses the web as the supervisor to generate annotations~\cite{hedderich2018training,mahajan2018exploring,sun2017revisiting}, since there is a large amount of imagery data with labeled information available on the web, such as social media hashtags and connections between web-pages and user feedback. 
%前者利用网络作为监督的来源，利用transfor learning等方法自动标记。后者更受到关注，因为所有可用的标签都是clean的,不会在标注上引入影响performance的噪音，另外数据集有很多可用，众包平台xxx. As shown in Table 1., Missing label means only a partial subset of full labels set can be observed: 我们只能知道图片中是否包含飞机，但不知道是否有汽车，轮船等对象
However, these methods may introduce additional noises to the label space, which can degrade a classifier’s performance.  
%To address this issue, some approaches \cite{kim2019nlnl,xiao2015learning} have been proposed to train multi-label classifier with noisy labels.  
%{\color{blue}{What are the restrictions or weakness of this approach?}}
%解决方案：train with missing label
For the latter, missing labels means that only a subset of all the labels can be observed and the rest remains unknown.  It can be further divided into several representative settings: fully observed labels (FOL), partially observed labels (POL), which is the most common setting.  Two variations of POL include: partially observed positive labels (PPL) and single positive label (SPL).
Table~\ref{table1} shows the difference between these settings.
It should be pointed out that POL setting is more common than PPL in real life.
For example, in many execution records of industrial devices~\cite{smith2015rolling, bin2012model, lee2016convolutional}, the probability of each component's failure is extremely low. Therefore, it is almost impossible to guarantee that each instance corresponds to one positive label, let alone in the setting of missing labels. 
%In this case, since the full label information is not required, the acquisition of datasets becomes easy. In addition, all available labels are accuracy and clean in this case, which simplifies the design of the structures of networks and complexity of implementation. 

\begin{table}[!tbp]
\caption{Different Missing-label Settings. \checkmark, $\times$, $\varnothing$ represent that current instance belongs to this class, does not belong to this class, and lacks related label, respectively.}
\begin{center}
\begin{tabular}{|c|c|c|c|c|c|}
\hline
\textbf{Settings}&\textbf{\textit{Class 1}}&\textbf{\textit{Class 2}}&\textbf{\textit{Class 3}}&\textbf{\textit{Class 4}}&\textbf{\textit{Class 5}} \\
\hline
FOL&\checkmark&$\times$&\checkmark &$\times$ &\checkmark  \\
\hline
POL&\checkmark&$\varnothing$&$\varnothing$ &$\times$ &\checkmark  \\
\hline
PPL&\checkmark&$\varnothing$&\checkmark &$\varnothing$ &$\varnothing$  \\
\hline
SPL&\checkmark&$\varnothing$&$\varnothing$ &$\varnothing$ &$\varnothing$  \\
\hline
\end{tabular}
\label{tab1}
\end{center}
\label{table1}
\end{table}

%xxxxx有待解决
This paper focuses on multi-label classification with missing labels.  Although there has been a lot of work done along this direction~\cite{zhang2021simple,wu2015ml}, there are still some critical issues to be~addressed: 
\begin{itemize}
    \item To solve the multi-label classification with missing labels, many state-of-the-art (SOTA) methods~\cite{durand2019learning}\cite{cole2021multi} rely on additional structures, such as GNN and label estimator, which further increase the complexity of networks. A natural question is whether this problem can be effectively solved without significantly increasing the network complexity.
\item It is still not clear how the missing ratio of the labels affects the classification performance, which is of great importance for us to balance the performance of classifier and the annotation cost.
\item Due to imbalance between positive and negative labels, most methods dealing with missing labels require that there is at least one positive label per instance, i.e., PPL, instead of POL, which is more common in real life. 
%并没有加入到讨论中在许多方法的设计时
\end{itemize}

%xx的方法仅通过改变loss来实现，但并没有对不同标签的缺失程度进行实验分析；  UPS等
%3.由于imbalance的存在，nagative标签的价值被严重低估。著名的pu learning类的方法只利用了positive标签的信息，而直接假设unlabel都是nagative的；oppo和cvpr21要求必须每个instance都必须有positive标签。而这种约束显然限制了现有方法在现实问题上的应用。

With these observations, this paper investigates new approaches for multi-label classification with missing labels.  The main contributions are summarized as follows:
\begin{itemize}
\item We propose a pseudo-label-based approach to predict all possible categories with missing labels, which can effectively balance the performance of classifiers and the cost of annotation. 
The network structure in our approach is the same as the classifier trained with full labels, which means that our approach will not increase the network complexity.  The major difference lies in the novel design of loss functions and training schemes.
\item We provide systematical and quantitative analysis of the impact of labels' missing ratio on the classifier's performance.  In particular, we relax the strict requirement that the label space of each instance must contain at least one positive label, which is  commonly seen in the related work~\cite{cole2021multi,zhang2021simple}.  Therefore, our method is applicable to general POL settings, not only PPL.
\item Comprehensive experiments verify that our approach can be effectively applied to missing-label classification.
Specifically, our approach outperforms most existing missing-label learning approaches, and in some cases even approaches trained with fully labeled datasets.  More importantly, our approach can adopt POL settings, which is incompatible with most existing methods.
%\item By quantifying the impact of negative labels on performance, we argue that ignoring negative labels in training due to the imbalance of positive and negative labels will squander valuable information of existing labels. Further, to alleviate the imbalance, we design a scheme by utilizing the proportion information of positives and negatives.
\end{itemize}

The rest of the paper is organized as follows.  Section~\ref{sec:related} discusses the related work.  The problem is formulated in Section~\ref{sec:form} and our proposed method is presented in Section~\ref{sec:PM}.  Section~\ref{sec:exp} shows the experimental results.  Finally, conclusions are drawn in Section~\ref{sec:con}.

%According to the number and positive/ negative properties of the observable labels, we divide multi-label problem into several settings:
%partially observed labels (POL), partially positive labels (PPL), partially negative labels (PNL) and two extreme cases, i.e. single positive label (SPL) and single negative label (SNL), as shown in Table.1.

%To our knowledge, this is the first work to systematically and quantitatively study all the above settings: because there exists an imbalance between the number of positives and negatives, which results in the domination of negatives on positives, \cite{cole2021multi}\cite{sun2010multi}\cite{hsieh2015pu}\cite{kanehira2016multi}\cite{han2018multi}\cite{su2021positive} mainly focus on positive labels and the value of negatives is severely underestimated; \cite{kim2019nlnl}\cite{duan2019learning}\cite{sun2017revisiting} use negative labels in training, but the contribution of negatives has not been quantified.

\section{Related Work}
\label{sec:related}

\subsection{Multi-label Learning with Missing Labels}
Recently, numerous methods have been proposed for multi-label classification with missing labels. Herein, we briefly review the relevant studies. 

\smallskip
\noindent
\textbf{\textit{Binary Relevance (BR).}} A straightforward approach for multi-label learning with missing labels is BR~\cite{tsoumakas2007multi,zhang2018binary}, which decomposes the task into a number of binary classification problems, each for one label. Such an approach encounters many difficulties, mainly due to ignoring correlations between labels. To address this issue, many correlation-enabling extensions to binary relevance have been proposed~\cite{cabral2011matrix,wu2015ml,xu2013speedup,yang2016improving,chen2013fast}. However, most of these methods require solving an optimization problem while keeping the training set in memory at the same time.  So it is extremely hard, if not impossible, to apply a mini-batch strategy to fine-tune the model~\cite{durand2019learning}, which will limit the use of pre-trained neural networks (NN)~\cite{kornblith2019better}.

\smallskip
\noindent
\textbf{\textit{Positive and Unlabeled Learning (PU-learning).}} PU-learning is an alternative solution~\cite{li2003learning}, which studies the problem with a small number of positive examples and a large number of unlabeled examples for training. Most methods can be divided into the following three categories: two-step techniques~\cite{liu2002partially,li2007learning,yu2002pebl}, biased learning~\cite{sellamanickam2011pairwise,claesen2015robust}, and class prior incorporation~\cite{jiang2008novel,liang2012learning}. All these methods require that the training data consists of positive and unlabeled examples \cite{bekker2020learning}.
In other words, they treat the negative labels as unlabeled, 
which discard the existing negatives and does not make full use of existing labels.
%and they discard the existing negative labels 

\smallskip
\noindent
\textbf{\textit{Pseudo Label.}} Pseudo-labeling was first proposed in \cite{lee2013pseudo}.  The goal of pseudo-labeling is to generate pseudo-labels for unlabeled samples~\cite{shi2018transductive}. There are different methods to generate pseudo labels: the work in~\cite{lee2013pseudo,rizve2021defense} uses the predictions of a trained NN to assign pseudo labels. Neighborhood graphs are used in~\cite{iscen2019label}. The approach in~\cite{wang2020repetitive} updates pseudo labels through an optimization framework. It is worth mentioning that MixMatch-family semi-supervised learning methods \cite{berthelot2019mixmatch,berthelot2019remixmatch,sohn2020fixmatch,xie2020unsupervised} achieve SOTA on multi-class problem by utilizing pseudo labels and consistency regularization~\cite{zhang2019consistency}. However, creation of negative pseudo-labels (i.e. labels which specify the absence of specific classes) is not supported by these methods, which therefore affects the performance of classifier by neglecting negative labels~\cite{rizve2021defense}. Instead, the work in~\cite{rizve2021defense} obtains the reference values of pseudo labels directly from the network predictions and then generates hard pseudo labels by setting confidence thresholds for positive and negative labels, respectively. Different from~\cite{rizve2021defense}, we simplify this process by studying the proportion of positive and negative labels to generate pseudo labels.

\subsection{Imbalance}
A key characteristic of multi-label classification is the inherent positive-negative imbalance created when the overall number of labels is large~\cite{ridnik2021asymmetric}. Missing labels exacerbate the imbalance and plague recognizing positives~\cite{zhang2021simple}. Therefore, the work in~\cite{zhang2021simple,cole2021multi} mandates that each instance in the training set must have at least one positive label, which means that they focus on PPL setting instead of ``real'' POL. Obviously, this assumption may not always hold in real life scenarios. To relax this assumption, a trivial solution is to treat the instances with only negative labels as unlabeled instances. In this case, however, it wastes the value of negative labels.

In this work, we allow the instances in training sets with only negative labels (that is POL setting).  From this perspective, our work is most closely related to~\cite{durand2019learning}. It should be pointed out that \cite{durand2019learning} uses a Graph Neural Network (GNN) \cite{scarselli2008graph} on top of a Convolutional Neural Network~(CNN) to model the correlations between labels, and its message update function relies on a multi-layer perception (MLP), which brings extra additional complexity to the parameter space and training process. Whereas our approach focuses on designing the training schemes and the loss function without introducing additional structures. Furthermore, although~\cite{durand2019learning} can cope with instances containing only negative labels, it does not specifically explore this direction, which leaves the possibility of falling into a trivial solution (always predict negative) when these types of instances occupies the majority. 

\section{Formulation}
\label{sec:form}
\subsection{Multi-label Classification}
Given a multi-label classifier with full labels, let $\mathcal{X}=\mathbb{R}^M$ be the input attribute space of $M$-dimensional feature vectors and $\mathcal{Y}=\{1, 2,\ldots,L\}$ denote the set of $L$ possible labels. 
An instance $\textbf{x}\in \mathcal{X}$ is associated with a subset of labels $\textbf{y}\in2^\mathcal{Y}$, which can be represented as an $L$-vector $\textbf{y}=[y_1,y_2,\ldots,y_L]=\{0,1\}^L$ where $y_j=1$ iff the $j$th label is relevant (otherwise $y_j=0$).
Let $\mathcal{D}=\{(\textbf{x}_1, \textbf{y}_1),\ldots,(\textbf{x}_N, \textbf{y}_N) \}$ is the training set of $N$ samples.
Given $\mathcal{D}$, a multi-label classifier $h:\mathcal{X}\rightarrow\mathcal{Y}$ learns to map the attribute input space to the label output space. We use $\hat{\textbf{y}}$ to present the prediction of classifier $h$, that is, $\hat{\textbf{y}}=f(h(\textbf{x}))$, where $f(\cdot)$ stands for a function (commonly the sigmoid function as $\sigma(s)=\frac{1}{1+e^{-s}}$) that turns confidence outputs into a prediction. In this case, most of the existing work \cite{nam2014large,tsoumakas2007multi,wang2016cnn,sorower2010literature} adopts the binary cross entropy (BCE) function as the loss function, which is formulated by
\begin{equation}
    \mathcal{L}(\hat{y},y)=-\frac{1}{L}\sum_{i=1}^{L}[(y_i \log(\hat{y}_i)+(1-y_i)\log(1-\hat{y}_i)]\label{bce}.
\end{equation}

For multi-label classification with missing-label, we use $\textbf{z}=[z_1,z_2,\ldots,z_L]=\{0,1,\varnothing \}^L$ to present the observed labels, where $z_j=\varnothing$ means the corresponding label is missing. 
It is worth mentioning that the training set is $\mathcal{D}=\{(\textbf{x}_1, \textbf{z}_1),\ldots,(\textbf{x}_N, \textbf{z}_N) \}$, while validation and test set is $\mathcal{D}_v=\mathcal{D}_t=\{(\textbf{x}_1, \textbf{y}_1),\ldots,(\textbf{x}_N, \textbf{y}_N) \}$ in this case. In other words, we still use full labels for validation and testing.

\subsection{Different Missing-label Settings}
According to the number and positive/negative properties of the observable labels, we divide multi-label classification problem into several settings:
partially observed labels (POL), partially positive labels (PPL), and an extreme case, i.e., single positive label (SPL) \cite{cole2021multi}. Specially, we formulate these settings as the following:
\begin{equation}
\begin{aligned}
    &\textbf{z}_{POL}=\{0,1,\varnothing\}^L \text{ and }\\ &\sum_{j=1}^L(\mathbf{1}_{[\textbf{z}_{POL_j}=1]}+\mathbf{1}_{[\textbf{z}_{POL_j}=0]})< L,
\end{aligned}
\end{equation}
\begin{equation}
\textbf{z}_{PPL}=\{1,\varnothing\}^L \text{ and } \sum_{j=1}^L\mathbf{1}_{[\textbf{z}_{PPL_j}=1]}< L,
\end{equation}
\begin{equation}
\textbf{z}_{SPL}=\{1,\varnothing\}^L \text{ and } \sum_{j=1}^L\mathbf{1}_{[\textbf{z}_{PPL_j}=1]}=1,
\end{equation}
%\begin{equation}
%\textbf{z}_{SNL}=\{0,\varnothing\}^L \text{ and } \sum_{j=1}^L\mathbf{1}_{[\textbf{z}_{PNL_j}=0]}=1,
%\end{equation}
where $\mathbf{1}_{[\cdot]}$ stands for the indicator function, that is, $\mathbf{1}_{[True]}=1$ and $\mathbf{1}_{[False]}=0$. Considering the large number of unknown labels and the imbalance problem in these settings, we design a new loss function to handle them, instead of directly using BCE~\eqref{bce}.

\section{Proposed Method}
\label{sec:PM}
\begin{figure*}[ht]
  \centering
  \includegraphics[width=0.95\linewidth]{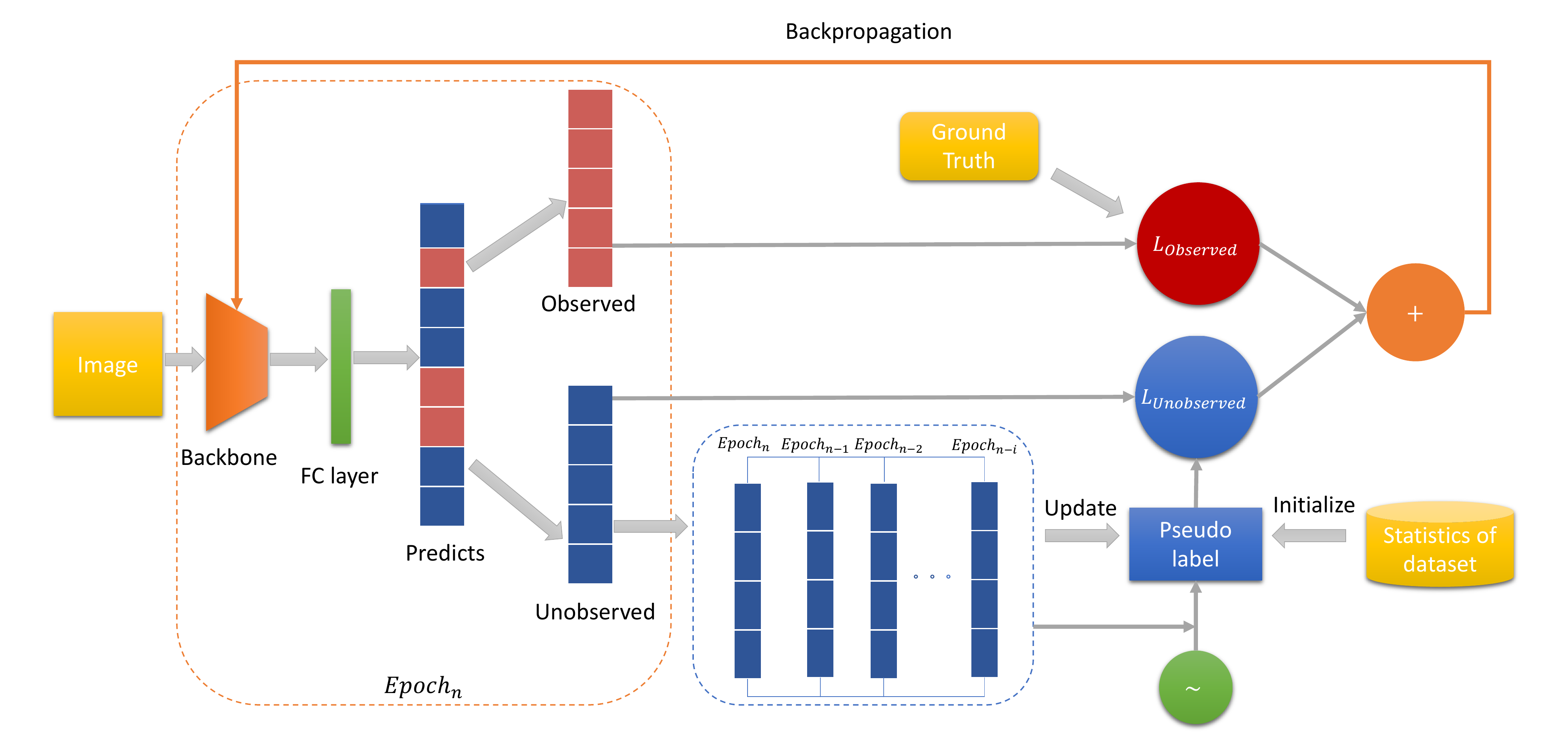}
  \caption{Pipeline of our method. We feed an image to the classifier network and obtain the predictions. Then different strategies are utilized to deal with observed and unobserved labels. For the unobserved part, we introduce pseudo labels by leveraging the prior knowledge of the dataset and design a novel loss function $\mathcal{L}_{\text{Unobserved}}$. For the observed part, we treat it as a full-label classification problem. The total loss is defined by adding the losses of these two parts with designed weights. The network used in our method is the most common classification network, which means that there is no additional complexity in the network structure. }
  %{\color{blue}{Make sure the figure is in the middle}} }
  \label{fig:framework}
\end{figure*}

Without changing the basic structure of the classification network, we address multi-label classification with missing labels by introducing pseudo-label, designing new loss functions, and adjusting training schemes. The pipeline of our method is shown as Fig.~\ref{fig:framework}.

\subsection{Pseudo-labels}
\label{subsec:pl}
For all missing-label settings, the label of any instance in the training set can be divided into two categories: observed and unobserved labels. 
\cite{sun2010multi,mac2019presence,bucak2011multi} directly set the unobserved labels as negative. 
However, performance drops because a lot of ground-truth positive labels are initialized as negatives \cite{joulin2016learning}. 
Therefore, we decide to introduce soft pseudo labels for the unobserved part and take pseudo labels as the target value of unobserved labels in calculation of the loss. 
Another benefit of introducing pseudo labels is that by properly designing updating methods, it will not add complexity to the existing classifier network.
In contrast, ROLE~\cite{cole2021multi} adds another NN as the label estimator and takes the predictions of this NN as the target. \cite{durand2019learning} adds a GNN on top of a CNN. Obviously, these methods bring additional complexity to the network.

In our approach, the classifier actually has two goals in the training process: 
\begin{itemize}
\item For observed labels, we expect the predicted value to be pushed to its ground-true value.
\item For unobserved labels, we expect the predicted value to be as close as possible to the value of its corresponding pseudo label.
\end{itemize}
All the design steps in our method revolve these two points. 
For (1), we just need to do as traditional supervised learning does.
For (2), there are two issues in front of us, that is, how to generate pseudo-labels and how to update them.
In the following discussion, we denote pseudo labels as $\widetilde{\textbf{y}}$ and the pseudo-labeling value corresponding to the $i$th instance's $j$th class as $\widetilde{y}_{i_j}$, which can be any real number between $[0,1]$.
%$ \widetilde{\textbf{y}}$来表示pseudo labels, y_i_j表示第i个实例的j类标签值。

\smallskip
\noindent
\textbf{\textit{Creation of Pseudo-labels.}} There are several approaches to create pseudo-labels for unobserved labels, as described in Section~\ref{sec:related}. 
In this work, we simplify this step by leveraging the prior knowledge of the datasets. 
We first calculate the average number of observed positive labels per instance and denote it as $P_p$, and then count the missing ratio $m$ of the current training set, $m=T_o/T$, where $T_o$ and $T$ stand for the number of observed and total labels, respectively. We expect that $P_p/m$ can, to some extent, approximate or recover the average number of positive labels per instance over the entire dataset.
With these statistics about the missing-label dataset, we can set the initial value for pseudo labels of the $i$th instance $\widetilde{y}_{i_j}$ as
\begin{equation}
    \widetilde{y}_{i_j} = 
    \begin{cases}
    \min\left(\frac{\max(\frac{P_p}{m},1) - P_i}{T_{u_i}},1\right), & \max(\frac{P_p}{m},1) > P_i\\
    0, & \textit{otherwise}
    \end{cases} \label{intipse}
\end{equation}
for unobserved label $j$, where $P_i$ and $T_{u_i}$ stand for the number of the observed positive labels and unobserved labels
in the $i$th instance, respectively. By using the $\max$ function in~\eqref{intipse}, it means we have an initial guess that each instance has at least one positive label in average, either observed or unobserved (of course, such an initial guess is not necessarily true). %{\color{blue}{this recovers PPL?}}

The initial method relies on a fact that even though most of the labels are missing, the statistics of the observed labels can still be used as a reference for the overall distribution of this dataset. Therefore, if the number of an instance's observed positive labels, $P_i$, is less than the average $\max(P_p/m,1)$, we assign the difference, divided by $T_{u_i}$, as the pseudo labels of those unobserved labels on this instance. Otherwise, if the number of the observed positive labels exceeds the average, we set the pseudo labels to 0. Note that this setting is only for initialization.  The pseudo labels will be updated during the training progresses.

It is worth mentioning that the previous approaches often initialize the pseudo labels as 0.5 or 0. Compared with the 0.5 setting, they do not take full advantage of the statistic information of the dataset. As to the 0 setting, it may aggravate the label imbalance.

\begin{figure}[ht]
  \centering
   \includegraphics[width=0.48\textwidth]{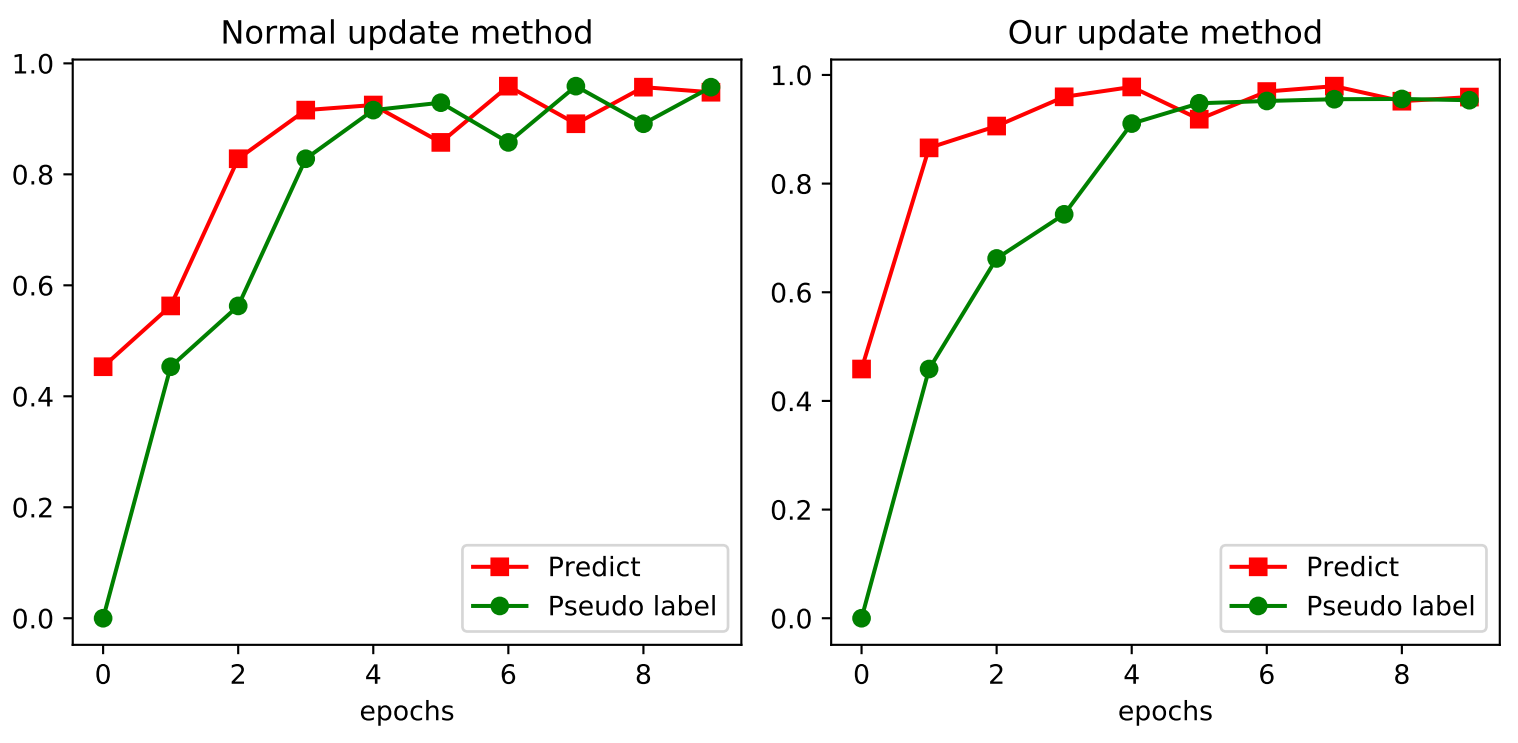}
   \caption{Fluctuations in predicted values and pseudo labels during training. The GT value for this specific label is 1. The left and right plot shows the trajectories of the predicted and pseudo labels under traditional updating method (set the pseudo label equal to the predicted value) and our running-average updating method, respectively. It is easy to see that our method accelerates convergence and smooths updating trajectories.}
	\label{fig:compare}
\end{figure}

\smallskip
\noindent
\textbf{\textit{Update.}}
We update the value of pseudo labels after each epoch of updating the network.  A common way is to simply use the network predictions as the pseudo labels.  To check the trajectory of the pseudo labels of this method,
we randomly select 10 training instances and record the fluctuations of their predictions and pseudo labels' value in SPL settings. 
%For comparison, we also record the GT values and the fluctuations of these 10 instances' predictions in FOL setting. 
%Due to space limitation, we only show the fluctuations of one category of one instance in Fig.~\ref{fig:compare} (the left and middle plots).
Due to space limitation, we only show the fluctuations of one category of one instance in Fig.~\ref{fig:compare}.
Obviously, the predictions fluctuate greatly in this common update method (as shown in the left plot in Fig.~\ref{fig:compare}). Note that even in the later stage of training, the fluctuation still exists, which makes it hard to converge. 
This is clearly not what we expect in training.

Therefore, we use a stack to record the predictions of our classifier for the past $n$ epochs and take the average of these historical predictions as the value for the associated pseudo labels, where $n$ is a hyperparameter.  Following this method, we record the trajectories of the predictions and the pseudo labels on the same training instances under SPL setting during training (the right plot in Fig.~\ref{fig:compare}). 
It is easy to find that this running-average method smooths the fluctuation of predictions and accelerate the convergence.
We formulate this updating method as follows:
\begin{equation}
    \widetilde{y}_{i_j} = \frac{1}{n} \sum_{k=0}^{n-1}\hat{y}_{i_j}^{m-k} \label{update}
\end{equation}
where $m$ stands for the current epoch index and $\hat{y}_{i_j}^{m}$ stands for the predicted value of the $j$th category in the $i$th instance at the $m$th epoch.

\smallskip
\noindent
\textbf{\textit{Disturbance Injection.}}
In experiments, we find that the predictions of the classifier for some unobserved categories remain around 0.5. 
As mentioned before, we believe that it is always better for the classifier to give an answer instead of being ambiguities. Following this concept, we devise a detector to detect such vague decisions.
By utilizing the historical prediction stack $S_{pse} = \{\hat{y}_{i_j}^{m-k}\}_{k=0}^{n-1}$, we determine whether these predictions are all in interval $[0.5-d, 0.5+d]$, where $d$ is a hyperparameter. If so, we update the pseudo labels using~\eqref{dis}, instead of updating pseudo labels in~\eqref{update},
which adds random disturbances to these kind of unobserved categories to push the pseudo labels away from 0.5:
\begin{equation}
    \widetilde{y}_{i_j} = 
    \begin{cases}
    \hat{y}_{i_j}- \text{random}(0, \hat{y}_{i_j}), & \hat{y}_{i_j} < 0.5\\
    \hat{y}_{i_j} + \text{random}(0,1-\hat{y}_{i_j}), & \text{otherwise}
    \end{cases}
    \label{dis}
    % if current value < .5, cv - random(0, current value)
    %if current >0.5, current + random(1-cv, 1)
\end{equation}
where $\text{random}(a,b)$ stands for a function which generates a random real number between $a$ and $b$.

Algorithm 1 summarizes the updating process of pseudo labels.  It should be pointed that the detection will not be performed in the first $D_s$ epochs and the last $D_e$ epochs of training, since the predicted labels may fall into $[0.5-d, 0.5+d]$ simply due to non-convergence at the beginning of training and training may stop before reaching convergence in the last several epochs. Here $D_s$ and $D_e$ are also hyperparameters.

\begin{algorithm}
\label{alg0}
	%\textsl{}\setstretch{1.8}
	\renewcommand{\algorithmicrequire}{\textbf{Input:}}
	\renewcommand{\algorithmicensure}{\textbf{Output:}}
	\caption{Update\_pseudo()}
	\label{alg1}
	\begin{algorithmic}[1]
	\REQUIRE predcited value $\hat{y_{i_j}}$ for $j$th class of $i$th instance at epoch $m$, stack $\textbf{S}_{i_j}$, start epoch $D_s$, end epoch $D_e$
	\IF{$D_s < m < D_e$ }
	\IF{$\textbf{S}_{i_j}$ is \textit{FULL}}
	\STATE $\textbf{S}_{i_j}$.\textit{pop}()
	\ENDIF
	\STATE $\textbf{S}_{i_j}$.\textit{push}($\hat{y_{i_j}}$)
	
	\IF{\textit{detector}($\textbf{S}_{i_j}$) is \textit{TRUE}}
	\STATE update pseudo labels $\widetilde{y}_{i_j}$ based on \eqref{dis}
	\ELSE
	\STATE $\widetilde{y}_{i_j}\leftarrow$ \textit{avg}($\textbf{S}_{i_j}$) based on \eqref{update}
	\ENDIF
	\ENDIF
	\ENSURE  $\widetilde{y}_{i_j}$, $\textbf{S}_{i_j}$
	\end{algorithmic}  
\end{algorithm}

%%%%%%%%%%%%%%%%%%%%%%%%%%%%%%%%%%%%%%%%%%%%%%%%%

\subsection{Design of Loss function}
At this point, we have solutions to meet the two goals mentioned in Subsection~\ref{subsec:pl}. Next, we relate these two goals together by designing an appropriate loss function. Intuitively, we can obtain the loss function by adding the losses of the observed part and unobserved part:
\begin{equation}
    \mathcal{L} = \mathcal{L}_{\text{Observed}} +  \mathcal{L}_{\text{Unobserved}}\label{bcepse}
\end{equation}
where $\mathcal{L}_{\text{Observed}}$ is obtained by modifying the number of classes $L$ in \eqref{bce} to the number of observed labels $O_i$ in the $i$th instance:
\begin{equation}
    \mathcal{L}_{\text{Observed}}=-\frac{1}{O_i}\sum_{j=1}^{O_i}[y_j\log(\hat{y}_j)+(1-y_j)\log(1-\hat{y}_j)].\label{ob}
\end{equation}
%we also need to modify the coefficients as what we did in \eqref{ob}. 
Before passing the pseudo labels into the loss function, we define a threshold function on the pseudo labels, $f(\widetilde{y_i},t)$, where $t$ is a hyperparameter. In order to simplify the description, this threshold is ignored below. 
%意义解释？？数学上
For $\mathcal{L}_{\text{Unobserved}}$, since the pseudo labels may contain errors, we can generally regard them as the GT values with noises. Therefore, we follow the idea in~\cite{wang2019symmetric} to design $\mathcal{L}_{\text{SBCE}}$ as follows, which has been proven to be robust to noises:
\begin{equation}
    \mathcal{L}_{\text{Unobserved}}=\mathcal{L}_{\text{SBCE}}=\alpha\mathcal{L}(\hat{y},\widetilde{y}) + \beta\mathcal{L}(\widetilde{y},\hat{y}).\label{SBCE}
\end{equation}
where $\alpha$ and $\beta$ stand for two decoupled hyperparameters. 

\smallskip
\noindent
\textbf{\textit{Imbalance between Positives and Negatives.}} Label imbalance significantly affects the generalization performance of the multi-label predictive model. The classifier can be easily reduced to the trivial ``always predict positive/negative'' solution. To alleviate imbalance, we add confidence scores %for the internals of $\mathcal{L}_{\text{Observed}}$ and
into $\mathcal{L}_{\text{Unobserved}}$ to balance the contributions of positive and negatives during
training: 
\begin{equation}
\begin{aligned}
        \mathcal{L}_{\text{CFS}}(x,y,C_1,C_2,L)=&-\frac{1}{L}\sum_{i=1}^{L}[C_1 x_i \log(y_i)\\&+C_2(1-x_i)\log(1-y_i)],
    \label{lcfs}
\end{aligned}
\end{equation}
where $C_1=\frac{N}{T}$ and $C_2=\frac{P}{T}$ are the weights, $N$ and $P$ stand for the total number of the observed negative and positive labels, respectively, and $T = N+P$ stands for the total number of observed labels in the training set. Obviously, $C_1+C_2=1$. Accordingly, the loss function for unobserved part can be defined as follows:
\begin{equation}
\begin{aligned}
     \mathcal{L}_{\text{Unobserved}}=&\alpha\mathcal{L}_{\text{CFS}}(\hat{y},\widetilde{y},C_1,C_2,U)\\& + \beta\mathcal{L}_{\text{CFS}}(\widetilde{y},\hat{y},C_1,C_2,U),\label{un}
     \end{aligned}
\end{equation}
where $U$ stands for the number of the unobserved labels on the current instance.
It should be pointed out that the focal loss in~\cite{lin2017focal} also designs scores for positive and negative labels, which are fine-tuned hyperparameters. Compared with~\cite{lin2017focal}, our method is more intuitive by using the scale information of observed labels in the training set.

%%%%%%%%%%%%%%%%%%%%%%%%%%%%%%%%%%%%%%%%%%%%%%%%%%%%%%%%%

\smallskip
\noindent
\textbf{\textit{The Weighted Loss Function.}}
To improve the loss, we follow the concept from curriculum learning~\cite{morerio2017curriculum}: start learning from the axioms that have been carefully tested, and then apply the axioms to real problems in life.  Consequently, 
we expect our model to follow the same pattern during the training process, i.e., it should first focus on the observed part, which takes the GT labels as target, and then gradually shifts its attention to the unobserved part.
Therefore, we design the time-varying confidence scores for $\mathcal{L}_{\text{Observed}}$ and $\mathcal{L}_{\text{Unobserved}}$ to reflect this dynamic by utilizing the index of the current epoch~$e$:
\begin{equation}
    \mathcal{L} = \frac{T_e-\frac{e}{2}}{T_e}\mathcal{L}_{\text{Observed}} + \frac{\frac{e}{2}}{T_e} \mathcal{L}_{\text{Unobserved}}.\label{fina}
\end{equation}
where $T_e$ is the total number of epochs.  We will use this loss function in~\eqref{fina} in our experiments.

%Different timed schedulers have been developed such as dropout training~\cite{morerio2017curriculum}, transfer learning~\cite{weinshall2018curriculum}, and self-correction~\cite{li2020self}

%分类器对预测为1往往是保守却可信的，我们相当于在初始阶段给分类器一个趋向于1的助力。
Algorithm 2 summarizes the entire training process of our approach for each instance.
\begin{algorithm}
	%\textsl{}\setstretch{1.8}
	\renewcommand{\algorithmicrequire}{\textbf{Input:}}
	\renewcommand{\algorithmicensure}{\textbf{Output:}}
	\caption{Training Process}
	\label{alg2}
	\begin{algorithmic}[1]
	\REQUIRE instance $\textbf{x}_{i}$, classifier $M(\cdot|\theta)$, ground-truth labels $G_i$, epoch index $m$, stack of $i$th instance $\textbf{S}_i$, pseudo labels of $i$th instance $\widetilde{y}_i$
	\IF{$\widetilde{y}_i$ not exists}
		\STATE $\widetilde{y}_i\leftarrow~$\textit{init}() based on \eqref{intipse}
		\ENDIF
		\STATE $\hat{y}_i\leftarrow M(\textbf{x}_{i}|\theta)$
		\STATE calculate $\mathcal{L}_{observed} (\hat{y}_i, G_i)$ based on \eqref{ob}
		\STATE calculate $\mathcal{L}_{unobserved} (\hat{y}_i,\widetilde{y}_i)$ based on \eqref{un}
		\STATE calculate $\mathcal{L}$ based on \eqref{fina}
		\STATE $\mathcal{L}$.\textit{backpropagation}()
		\STATE update parameter $\theta$
		\REPEAT
		\STATE $\hat{y}_{i_j}$, $\textbf{S}_{i_j}\leftarrow~$Update\_pseudo($\hat{y}_{i_j}$, $\textbf{S}_{i_j}$)
		\UNTIL perform the above for all unobserved classes in $\textbf{x}_{i}$
		\STATE $m\leftarrow m+1$
		\ENSURE $M(\cdot|\theta)$, $m$
	\end{algorithmic}  
\end{algorithm}

\section{Experiments}
\label{sec:exp}
%different settings, different dataset
We test the effectiveness of our approach separately on different labeling settings (POL, PPL, and SPL) and various datasets, and compare it with several representative baseline methods. Then we provide the ablation study that evaluates the contribution of each component described in Section~\ref{sec:PM}.

\subsection{Datasets}
%介绍如何
%数据集介绍
%标签筛选
%训练测试集划分
%end to end
We conduct comprehensive experiments on three large-scale multi-label image datasets: COCO~\cite{coco}, NUS-WIDE~\cite{nus}, and Pascal VOC~\cite{pascal}.  Each instance in these three datasets is fully annotated with clean labels that can be used as the GT in performance evaluation. 

To test our approach on various missing-label settings, we need to corrupt label spaces of these three datasets by discarding a part of labels. The processing method we use is similar to~\cite{cole2021multi} which generate labels for SPL setting.  The difference in our case is that besides SPL setting, we also generate other settings by randomly selecting positive/negative labels at specified proportions. %Specifically, we randomly select a part of labels in each instance in training set and retain them, 
If a label (0 or 1) is discarded, we use ${\varnothing}$ to denote it. Note that we only do this operation once so that the labels on each instance of the training set can keep consistent during multiple training processes. 

For COCO and Pascal VOC, we adopt the official training/testing splitting methods. 
%See supplementary for details. 
In COCO, the official training set consists of 82081 images, where each one is associated with 80 possible categories, and there are 40,137 images in the official testing set. 
The official training and testing sets of Pascal VOC contain 5,717 and 5,823 images respectively, and each image has 20 different categories. 
For NUS-WIDE, we download it from Flickr, mix the official training and testing sets together, and then follow the splitting method of \cite{ridnik2021asymmetric}. Finally, we collect NUS-wide with 116445 and 50720 images as training set and test set and each image has 81 labels.

After determining the training and testing sets, to be consistent with~\cite{cole2021multi}, we choose \textbf{20\%} images from the training set as the validation set.
Then we train end-to-end fine-tuned deep CNNs across these three datasets in different settings.

It should be pointed out that since we adopt the round-up method during the process of randomly selecting labels, PPL\_08 for Pascal VOC is equivalent to using all positive labels while none negative labels are selected.
Table~\ref{table2} shows the statistics of positive and negative labels used in different settings and different datasets.

As for the data augmentation and pre-processing, we first resize the original input image of all these three datasets to the shape of $448\times 448$. And then, horizontal flip is applied with a probability of 0.5. In the end, by using the standard Imagenet statistics, we normalize the input image. 

\begin{table}[!htbp]
\begin{center}
\caption{Statistics of the observed labels in different settings mentioned in this work, including the total number of positive/negative labels and the average number of positive/ negatives labels per instance. We bold the number of positive labels per instance less then 1, which proves that there exists some instances with \textbf{no} positive labels in these settings. This also demonstrate that our method relax the requirement that each instance must contain at least one positive label.}
%{\color{blue}{This is not consistent with the number of images in each dataset.  Please check the numbers, especially for SPL.}} 
\begin{tabular}{|l|l|l|l|l|}

\hline

        & \multicolumn{1}{l|}{total pos} & \multicolumn{1}{l|}{per. pos} & \multicolumn{1}{l|}{total neg} & per. neg \\ \hline
Dataset                & \multicolumn{4}{c|}{Pascal VOC}                                                                            \\ \hline
FOL     & \multicolumn{1}{l|}{6665}          & \multicolumn{1}{l|}{1.5}          & \multicolumn{1}{l|}{84815}         & 18.5         \\ \hline
PPL\_08 & \multicolumn{1}{l|}{6665}          & \multicolumn{1}{l|}{1.5}          & \multicolumn{1}{l|}{0}         &       0   \\ \hline
PPL\_06 & \multicolumn{1}{l|}{6244}          & \multicolumn{1}{l|}{1.4}          & \multicolumn{1}{l|}{0}         &    0      \\ \hline
PPL\_04 & \multicolumn{1}{l|}{5009}          & \multicolumn{1}{l|}{1.1}          & \multicolumn{1}{l|}{0}         &      0    \\ \hline
SPL     & \multicolumn{1}{l|}{4574}          & \multicolumn{1}{l|}{1.0}          & \multicolumn{1}{l|}{0}         &      0    \\ \hline
POL\_005 & \multicolumn{1}{l|}{340}          & \multicolumn{1}{l|}{\textbf{0.1}}          & \multicolumn{1}{l|}{4234}         &      0.9    \\ \hline
POL\_01 & \multicolumn{1}{l|}{649}          & \multicolumn{1}{l|}{\textbf{0.1}}          & \multicolumn{1}{l|}{8499}         &       1.9   \\ \hline
POL\_02 & \multicolumn{1}{l|}{1302}          & \multicolumn{1}{l|}{\textbf{0.3}}          & \multicolumn{1}{l|}{16994}         &      3.7    \\ \hline
POL\_04 & \multicolumn{1}{l|}{2662}          & \multicolumn{1}{l|}{\textbf{0.6}}          & \multicolumn{1}{l|}{33930}         &     7.4     \\ \hline
POL\_06 & \multicolumn{1}{l|}{3974}          & \multicolumn{1}{l|}{\textbf{0.9}}          & \multicolumn{1}{l|}{50914}         &     11.1     \\ \hline
POL\_08 & \multicolumn{1}{l|}{5315}          & \multicolumn{1}{l|}{1.2}          & \multicolumn{1}{l|}{67869}         &     14.8     \\ \hline
   Dataset     & \multicolumn{4}{c|}{COCO}                                                                                  \\ \hline
FOL     & \multicolumn{1}{l|}{190378}          & \multicolumn{1}{l|}{2.9}          & \multicolumn{1}{l|}{5060122}         &         77.1 \\ \hline
PPL\_08 & \multicolumn{1}{l|}{186723}          & \multicolumn{1}{l|}{2.8}          & \multicolumn{1}{l|}{0}         &    0      \\ \hline
PPL\_06 & \multicolumn{1}{l|}{152609}          & \multicolumn{1}{l|}{2.3}          & \multicolumn{1}{l|}{0}         &    0      \\ \hline
PPL\_04 & \multicolumn{1}{l|}{111160}          & \multicolumn{1}{l|}{1.7}          & \multicolumn{1}{l|}{0}         &    0      \\ \hline
SPL     & \multicolumn{1}{l|}{65665}          & \multicolumn{1}{l|}{1.0}          & \multicolumn{1}{l|}{0}         &     0     \\ \hline
   Dataset     & \multicolumn{4}{c|}{NUS-WIDE}                                                                              \\ \hline
FOL     & \multicolumn{1}{l|}{176998}          & \multicolumn{1}{l|}{1.9}          & \multicolumn{1}{l|}{7368638}         &     79.1     \\ \hline
PPL\_08 & \multicolumn{1}{l|}{166818}          & \multicolumn{1}{l|}{1.8}          & \multicolumn{1}{l|}{0}         &     0    \\ \hline
PPL\_06 & \multicolumn{1}{l|}{142734}          & \multicolumn{1}{l|}{1.5}          & \multicolumn{1}{l|}{0}         &    0      \\ \hline
PPL\_04 & \multicolumn{1}{l|}{114581}          & \multicolumn{1}{l|}{1.2}          & \multicolumn{1}{l|}{0}         &  0        \\ \hline
SPL     & \multicolumn{1}{l|}{93156}          & \multicolumn{1}{l|}{1.0}          & \multicolumn{1}{l|}{0}         &      0    \\ \hline
\end{tabular}
\label{table2}
\end{center}
\end{table}

\subsection{Metrics}
There are different standard metrics to evaluate the performance of multi-label classifiers~\cite{tsoumakas2007multi,sorower2010literature}, such as mean average precision(mAP), precision-at-k, and recall-at-k, to name a few. In our case, we adopt mAP as the primary evaluation metric.

\subsection{Baselines}
\label{bl}
In order to demonstrate the effectiveness of our method and evaluate the impact of the missing ratio, we first choose \textbf{BCE} and \textbf{BCE-LS} with full labels as strong baselines. 
These two methods are the most commonly used methods in multi-label problem. The former takes BCE~\eqref{bce} as the loss function and the latter uses label smoothing BCE~\cite{szegedy2016rethinking}, which is proposed to reduce overfitting and has been shown to be effective in mitigating the negative impacts of label noises.

Besides taking \textbf{BCE} and \textbf{BCE-LS} under FOL setting as strong baselines, we select some representative approaches for comparison.

In PPL setting, we choose \textbf{AN}~\cite{kundu2020exploiting}, \textbf{WAN}~\cite{mac2019presence}, and \textbf{ROLE}~\cite{cole2021multi} and directly use the official experimental settings and code.
\textbf{AN} assumes that unobserved labels are always negative, which is perhaps the most common method for PPL settings.
\textbf{WAN} down-weights terms in the loss related to negative labels by introducing a weight parameter.
%$\gamma\in[0,1]$. Here we choose $\gamma =1/(L-1)$ in our experiments, where $L$ stands for the number of total categories in the dataset.
\textbf{ROLE} considers regularized online estimation of unobserved labels.

For POL, as stated before, most existing methods can not be applied in this setting directly. 
Therefore, we have to make some modifications to the original methods. 
Specifically, besides \textbf{AN}, \textbf{WAN} and \textbf{ROLE}, the approaches for comparison in this setting include \textbf{Focal}~\cite{lin2017focal}, and \textbf{ASL}~\cite{ridnik2021asymmetric}.
For \textbf{AN}, \textbf{WAN} and \textbf{ROLE}, due to the severe imbalance in missing-label settings, Cole et al.~\cite{cole2021multi} reuqire that each instance must have one positive labels in the code implementation.
We directly remove the restriction in the official code for comparison. 
\textbf{Focal} addresses the problem of positive-negative imbalance and hard-mining.  We can formulate it as follows:
\begin{equation}
-\sum_{i=1}^L(y_i\alpha_+(1-p)^{\gamma}\log(p) + (1-y_i)\alpha_-p^{\gamma} \log(1-p))
\end{equation}
where $p$ stands for the predict value, $\gamma$ is a focus parameter and $\alpha_+,\alpha_-$ are used to balance positive and negative labels. 
In our experiment, we set $\alpha_+=0.9$, $\alpha_-=0.1$ and $\gamma=2$, which is determined experimentally.  
\textbf{ASL} relieves imbalance by operating differently on positives and negatives as follows,
\begin{equation}
    -\sum_{i=1}^L(y_i(1-p_m)^{\gamma_+}\log(p_m) + (1-y_i)p_m^{\gamma_-} \log(1-p_m))\label{asl}
\end{equation}
where $\gamma_+,\gamma_-$ are the focus parameters, $p_m=\max(p-m,0)$, and $m$ is a hyperparameter. We set $\gamma_+=8$, $\gamma_-=1$ and $m=0.05$ for our experiments through fine tuning.
Note that although \textbf{Focal} and \textbf{ASL} are designed to solve the imbalance issue, the authors did not consider them in missing-label settings. Therefore, in order to make a comparison with our method, we follow~\cite{bucak2011multi, mac2019presence,sun2010multi} and regard the labels for unobserved part as negative when implementing these two approaches.

\begin{table}[tbp]
\caption{The mAP results in POL settings on Pascal VOC. In each column we bold the best performing method and underline the second-best except \textbf{BCE} under FOL setting.}
\begin{center}
\begin{tabular}{|l|l|l|l|l|}
\hline
& POL\_02 & POL\_04 & POL\_06 & POL\_08 \\ \hline
%%%%%%%%%%%%%%%%ours调C1，C2，可能会提升%%%%%%%%%%%%%%%%%%%%%%%%%%%
BCE (FOL) &       \multicolumn{4}{|c|}{89.1}       \\ \hline
Ours &    \textbf{81.5}    &   \textbf{86.5}    &      \textbf{88.0}   &     \textbf{89.2}    \\ \hline
ASL \cite{ridnik2021asymmetric} & \underline{78.2}       &    \underline{84.6}    &    \underline{86.7}     &    \underline{88.0}     \\ \hline
Focal \cite{lin2017focal} &    77.7     &    82.1     &    85.9     &      87.3   \\ \hline
AN~\cite{kundu2020exploiting} &   63.1    &   69.4    &   72.8    & 76.5      \\ \hline
WAN~\cite{mac2019presence} &   70.6    &   78.9    &   80.0    & 83.4      \\ \hline
ROLE~\cite{cole2021multi} &   69.2    &   79.0    &   81.1    & 82.9      \\ \hline

%%%%%%%focal算法有待调参，可能会提升%%%%%%%%%%%%%%%%%%%%%%

\end{tabular}
\end{center}
\label{table:POL}
\end{table}

%an wan IU role focal asl

\subsection{Implement Details}

\noindent
\textbf{\textit{Network Structure.}}
We use an end-to-end network for all experiments: a ResNet-50 \cite{xie2017aggregated}, pre-trained on ImageNet \cite{deng2009imagenet}, as the backbone and a fully connected layer, which is the same as the multi-label classifier under FOL setting. Our approach does not add any extra structure to the network.

\smallskip
\noindent
\textbf{\textit{Hyperparameters.}}
We train our classifier for 10 epochs, that is, $T_e = 10$. 
For the learning rate and batch size, we use a hyperparameter search method and select the hyperparameters with the bast mAP on the validation set, where the learning rate is in $[1e-3, 1e-4, 1e-5, 1e-6]$ and batch size is in $[8,16]$.
We assign the threshold $t$ to be $0.7$ and $d$ to be $0.2$.  Recall that $d$ is used to decide whether to introduce disturbances in pseudo labels.
The historical stack size $n$ is 3, considering the memory consumption.  $D_s$ and $D_e$ are set to be 3 and 7, respectively, which are determined after several trials.
Finally, we set $\alpha=0.95$ and $\beta=0.05$.
%equation6中的n；Ds，De；alpha beta
%%%%%%%%%%%%%%%%%%%%%%%%%%%%还在调参数，此处有待修改%%%%%%%%%%%%%%%%%%%%%%%%%%%%%%
\begin{figure}[ht]
  \centering
   \includegraphics[width=0.45\textwidth]{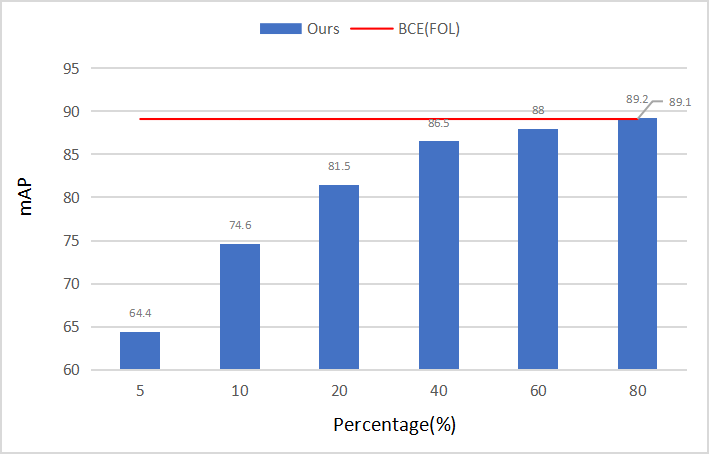}
    \caption{The performance of our approach in POL settings of Pascal VOC. The x-axis represents the percentage of observed labels used in our experiments. For instance, 5 stands for POL\_005. The y-axis represents the mAP scores. The score of \textbf{BCE} under FOL setting is shown in red for comparison. Notice that in POL\_08, our approach eventually \textbf{exceeds} the performance of \textbf{BCE} with full labels. }
	\label{fig:test}
\end{figure}
\subsection{Analysis}

%随着label数量变化的map折线图，positive label数量，nagetivelabel数量，总数量
%每种setting，绘制一个表格来比较
%pseudo label随着训练进行的变化情况 折线
%loss的变化情况随着训练 折线

%这组实验证明方法对应missing label的有效性

\noindent

\begin{table*}[htbp]
\label{table3}
\caption{The mAP results in PPL settings on COCO, Pascal VOC, and NUSWIDE datasets. PPL\_04, PPL\_06, and PPL\_08 represent randomly retaining 40\%, 60\% and 80\% of positive labels for each instance, respectively, and discarding all negative labels. SPL is equivalent to PPL\_single, which is an extreme variation of PPL. 
In each column we bold the best performing method and underline the second-best except \textbf{BCE} and \textbf{BCE-LS} that is performed under FOL setting.}
\begin{center}
\begin{tabular}{|c|c|c|c|c|c|c|c|c|c|c|c|c|}
\hline
\multirow{3}*{}&\multicolumn{4}{|c|}{\textbf{COCO}}&\multicolumn{4}{|c|}{\textbf{Pascal VOC}}&\multicolumn{4}{|c|}{\textbf{NUSWIDE}} \\
%\cline{2-10}
%&\multicolumn{9}{|c|}{\textbf{FOL}}\\
%\midrule[1.25pt]
\hline
 & SPL & PPL\_04 &PPL\_06& PPL\_08 & SPL & PPL\_04&PPL\_06 & PPL\_08 & SPL & PPL\_04 &PPL\_06& PPL\_08\\
\hline
BCE (FOL) & \multicolumn{4}{|c|}{75.8} &  \multicolumn{4}{|c|}{88.9} & \multicolumn{4}{|c|}{52.6} \\
\hline
BCE-LS (FOL) & \multicolumn{4}{|c|}{76.8} &  \multicolumn{4}{|c|}{90.0} & \multicolumn{4}{|c|}{53.5} \\
\hline
AN \cite{kundu2020exploiting} &63.9 &66.4& 67.0&69.2 &84.7 &86.8&87.3 &88.0 &40.0 &45.1& 48.3&50.3 \\
\hline
WAN \cite{mac2019presence}&64.8 & 69.1&70.8&71.2 &85.9 &87.5& 87.9&88.3 &43.7& 46.9&48.5&50.6 \\
\hline
ROLE \cite{cole2021multi} &\underline{65.9} & \underline{73.1}&\underline{75.4} &\underline{77.1} &\textbf{87.0} &\textbf{88.9}&\underline{90.0} &\underline{90.3} &\underline{43.2} &\underline{48.0} &\underline{50.4} & \underline{52.0}\\
\hline
Ours &\textbf{67.8} &\textbf{74.4}& \textbf{75.9}&\textbf{77.8} &\underline{86.6} &\textbf{88.9}& \textbf{90.8}&\textbf{91.1} &\textbf{46.0} &\textbf{49.6}&\textbf{51.9} &\textbf{53.3} \\
%\midrule[1.25pt]
\hline
\end{tabular}
\label{table3}
\end{center}
\end{table*}

\textbf{\textit{POL.}}
Alleviating the imbalance between positives and negatives is one highlight of our approach. 
In POL, the impact of the imbalance may be exacerbated since the observable labels are randomly selected.  There is equal chance for positive and negative labels to be selected, which means that it is even possible that an instance has multiple positive labels, while another one has no positive labels. This is the reason why most related methods are incompatible with this setting.
% Since \textbf{AN}, \textbf{WAN}, and \textbf{ROLE} stipulate that each instance must contain at least one positive label, these methods cannot be applied to this setting directly.
We opt to use the modified version of \textbf{AN}, \textbf{WAN}, \textbf{ROLE}, \textbf{Focal} and \textbf{ASL} as baseline methods for comparison, and the detailed modification process is described in Sec.~\ref{bl}.

We conduct experiments on Pascal VOC in POL\_02, POL\_04, POL\_06, and POL\_08 settings. 
% \textbf{BCE} has no requirement on the number of positive labels per instance, but this method can not be applied to missing-label settings directly. Therefore, we perform a variant of \textbf{BCE}, \textbf{BCE} by assmuing the missing part as negatives (\textbf{BCE-AN}), in the same settings. 
%{\color{blue}{Shouldn't BCE be under FOL?  Why still missing labels?  Any difference from PPL setting for BCE?}}
The experimental results are shown in Table~\ref{table:POL}.
As excepted, since \textbf{AN} does not take imbalance into consideration, it performs the worst among all other methods in our experiments.
Moreover, \textbf{ROLE}, which won the second place in PPL setting in our experiments, can not cope with POL setting well. 
Relying on the ingenious design of the loss function, our method has a significant improvements compared with other baseline-methods.
In particular, we notice that the fewer labels are available, the greater the performance of our approach can be over other baselines. 
Obviously, the experimental results show that our approach can effectively deal with POL settings, which is the most common settings in real world.

%However, due to the lack of design for the missing-label classification problems, we cannot achieve ideal performance by simply assuming missing part as negatives ~\cite{durand2019learning}. 
%%%%%%%%%%%%%%%%%%%%%POL 0.01， 0.05， 0.1， 0.2， 0.4， 0.6，0.8 our approach 画一个图%%%%%%%%%%%%%%
To further quantify the impact of the missing ratio of labels on the classification performance, we run our approach in POL\_005, POL\_01, and the results are summarized in Fig.~\ref{fig:test}. 
The performance of our algorithm improves as the number of observed labels increases. Note that in POL\_06, we use only 60\% of labels and the achieved score is only 1.1\% lower than the strong baseline. In POL\_08, the performance even exceeds the strong baseline with only 80\% of labels.
Considering that the cost of annotation is expensive, we can find a balance between 60\% and 80\% of labels to trade off accuracy of the classifier and the annotation cost.

%\textbf{\textit{Imbalance ratio.}}
%To further quantify The impact of labels’ missing ratio on performance
%remains to be quantified

%正标签的数量   正负标签的比例  哪个更重要
%正标签的数量   正负标签的比例  哪个更重要
%正标签的数量   正负标签的比例  哪个更重要
%正标签的数量   正负标签的比例  哪个更重要

\smallskip
\noindent
\textbf{\textit{PPL \& SPL.}}
In order to demonstrate the effectiveness of our approach in PPL settings, we consider different labeling settings (PPL\_04, PPL\_06, PPL\_08 and SPL) on three datasets.
For fair comparison, we directly execute the official code of \textbf{AN}, \textbf{WAN} and \textbf{ROLE} without any modification as baseline-methods.  
Meantime, we perform \textbf{BCE} and \textbf{BCE-LS} on these three fully-labeled datasets, i.e., under FOL setting, which are used as strong baselines for our approach.
Table~\ref{table3} summarizes the experimental results.
%
%
%
%%%%%%%%%%%%%%%%%%%%%%%%%%%%%%%新增了PPL06的实验，用于确定missing ratio多少能达到FOL精度%%%%%%%%%%%%%%%%%%%%
%%%%%%%%%%%%%%%%%%%%%%%%%%%%%%%5等实验结果%%%%%%%%%%%%%%%%%%%%%%%%%%%%%%%%%%%%%%%
%%%%%%%%%%%%%%%%%%%%%%%%%%%%%%%别忘了%%%%%%%%%%%%%%%%%%%%%%%%%%%%%%%%%%%%
For COCO dataset and NUS-WIDE dataset, our algorithm achieves the highest score in all settings. 
For Pascal VOC, our approach achieves the best mAP score in PPL\_04, PPL\_06 and PPL\_08 settings and the second highest, but comparable, score in SPL setting.   
These results fully demonstrate that our approach can be effectively applied to PPL settings.

Compared with the strong baselines under FOL setting, the results show that for COCO, our approach can achieve comparable performance to \textbf{BCE} and \textbf{BCE-LS} in PPL\_06 and PPL\_08, respectively.  
For Pascal VOC, the results by our approach in PPL\_06 and PPL\_08 are even able to outperform these two FOL baselines. 
It is surprising to find that our approach in PPL\_04 can even achieve a score comparable to strong baselines,
which means we save 26.7\% of the cost in labeling positives and 100\% cost in labeling negatives, according to the statistics of the labels used in our experiments.  
As for NUS-WIDE, our method also can reach the strong baselines ,even outperforms \textbf{BCE(FOL)} in PPL\_08 setting. However, the other baseline-methods are not able to exceed the strong baselines.

\subsection{Ablation study}
\label{subsec:AS}
We present an ablation study to measure the contributions of different components in our approach. We run experiments on Pascal VOC in PPL\_06 and POL\_06.  The results are shown in Table~\ref{table:AS}. 

\begin{table}[htbp]
\caption{The mAP results in ablation study under PPL\_06 and POL\_06 settings on Pascal VOC. }
\begin{center}
\begin{tabular}{|l|l|l|}
\hline
     & PPL\_06 & POL\_06\\ \hline
Ours Approach&     90.8    &88.0    \\ \hline
Without Update &90.1 &86.6 \\ \hline
Without Disturbances &88.2 &86.0 \\ \hline
Without Imbalance Design &91.1 & 76.9 \\ \hline
Without Weighted Loss &90.9 &79.4 \\ \hline
\end{tabular}
\label{table:AS}
\end{center}
\end{table}

For PPL\_06 setting, both our proposed running-average updating method and the introduction of disturbances improve the performance of the approach. However, %in order to be compatible with POL settings and alleviate the imbalance,
the novel loss function, including imbalance and weighted design, reduces the performance, because these two factors are designed for more general settings, not specifically for PPL.
This can be found in POL\_06 setting, where the effectiveness of our design is demonstrated: the imbalance and dynamically weighted design can significantly enhance the performance of the classifier.  The former improves the result by more than 11.1\%, and the latter improves by 8.6\%.
In contrast, the new updating method and disturbance have relatively limited, but still positive, improvements in POL setting (more than 1\% increase).

\section{Conclusion}
\label{sec:con}
This paper explores the problem of multi-label classification with missing labels.
We propose a novel pseudo-label-based approach to cope with missing-label problem without increasing the complexity of the existing classification networks.  By leveraging prior knowledge of the dataset, our approach relaxes the assumption that each instance in the training set must has at least one positive label, which is often required in existing approaches.  We demonstrate the effectiveness of our approach by performing extensive experiments on three large-scale datasets. It is shown that our method can effectively reduce the annotation cost without significant degradation in classification performance.  We also evaluate the impact of the missing ratio of the labels on the classifier's performance, which can be used as a guideline to achieve the balance between performance and labeling costs. 

%\textbf{\textit{Future. }} 
%negative标签的能力被极大的低估了，这仍然有很大的提升空间。large label space的解决方案(label space in NUS is worse then  coco,因为空间太大)，考虑到实际生活的多标签性，如果将现有方法应用在大标签空间上亟待解决。

\bibliographystyle{IEEEtran}
\bibliography{reference}

\end{document}